# Assessing the (Un)Trustworthiness of Saliency Maps for Localizing Abnormalities in Medical Imaging

Nishanth Arun[†1,2], Nathan Gaw[†3], Praveer Singh[1], Ken Chang[1,4], Mehak Aggarwal[1], Bryan Chen[1,4], Katharina Hoebel[1,4], Sharut Gupta[1], Jay Patel[1,4], Mishka Gidwani[1], Julius Adebayo[4], Matthew D. Li[1], and Jayashree Kalpathy-Cramer[*1]

[1]Athinoula A. Martinos Center for Biomedical Imaging, Department of Radiology, Massachusetts General Hospital, Harvard Medical School, Boston, MA, USA
jkalpathy-cramer@mgh.harvard.edu
https://qtim-lab.github.io/
[2]Shiv Nadar University, Greater Noida, India
[3]H. Milton Stewart School of Industrial and Systems Engineering, Georgia Institute of Technology, Atlanta, GA, USA
[4]Massachusetts Institute of Technology, Cambridge, MA, USA

[†]*These authors contributed equally*
[*]*Corresponding author*

**Purpose:** To evaluate the trustworthiness of saliency maps for abnormality localization in medical imaging.

**Materials and Methods:** Using two large publicly available radiology datasets (SIIM-ACR Pneumothorax Segmentation and RSNA Pneumonia Detection), we quantified the performance of eight commonly used saliency map techniques in regards to their 1) localization utility (segmentation and detection), 2) sensitivity to model weight randomization, 3) repeatability, and 4) reproducibility. We compared their performances versus baseline methods and localization network architectures, using area under the precision-recall curve (AUPRC) and structural similarity index (SSIM) as metrics.

**Results:** All eight saliency map techniques fail at least one of the criteria and were inferior in performance compared to localization networks. For pneumothorax segmentation, the AUPRC ranged from 0.024-0.224, while a U-Net achieved a significantly superior AUPRC of 0.404 ($p<0.005$). For pneumonia detection, the AUPRC ranged from 0.160-0.519, while a RetinaNet achieved a significantly superior AUPRC of 0.596 ($p<0.005$). Five and two saliency methods (out of eight) failed the model randomization test on the segmentation and detection datasets, respectively, suggesting that these methods are not sensitive to changes in model parameters. The repeatability and reproducibility of the majority of the saliency methods were worse than localization networks for both the segmentation and detection datasets.

**Conclusion:** We suggest that the use of saliency maps in the high-risk domain of medical imaging warrants additional scrutiny and recommend that detection or segmentation models be used if localization is the desired output of the network.

*Supplemental material is available for this article.*

**Summary:** The use of saliency maps to interpret deep neural networks trained on medical imaging fails several key criteria for utility and robustness, highlighting the need for scrutiny before clinical application.

**Key Points:**

1. Eight popular saliency map techniques were evaluated for their utility and robustness in interpreting deep neural networks trained on chest radiographs.

2. All the saliency map techniques fail at least one of the criteria defined in the paper, indicating their use for high-risk medical applications to be problematic.

3. Instead, the use of detection or segmentation models are recommended if localization is the ultimate goal of interpretation.

## Introduction

Deep learning has brought many promising applications within medical imaging with recent studies showing potential for key clinical assessments within radiology[1, 2, 35]. One major class of deep neural networks is convolutional neural networks (CNNs), which take raw pixel values as input, and transform them into the output of interest (such as diagnosis). Many CNNs have outperformed conventional methods for various medical tasks[26, 27]. As CNNs are becoming popular for classification of medical images, it has become important to find methods that explain the decisions of these models to establish trust with clinicians. Saliency maps have become a popular approach for post-hoc interpretability of CNNs. They are designed to highlight the salient components of medical images that are important to model prediction. As a result, many CNN medical imaging studies have used saliency maps to rationalize model prediction and provide localization[3, 4, 5]. However, a recent study that evaluated a variety of datasets showed that many popular saliency maps are not sensitive to model weight or label randomization[3]. Although there is no study that corroborates these findings with medical images, there are several works that have demonstrated serious issues with saliency methods[7, 5, 10]. A recent study also showed that saliency maps did not provide additional performance improvement in assisted clinician interpretation compared to only providing a model prediction[34]. To the best of our knowledge, [5] is the only work that has evaluated the robustness of saliency maps in medical imaging. However the work does not encompass all the widely used saliency methods nor effectively quantify the overlap of saliency maps with ground-truth regions.

In this study, we comprehensively evaluate popular saliency maps for CNNs trained on the SIIM-ACR Pneumothorax Segmentation and RSNA Pneumonia Detection datasets[11, 37, 12, 36] in terms of 4 key criteria for trustworthiness:

> 1. Utility
>
> 2. Sensitivity to weight randomization
>
> 3. Repeatability (intra-architecture)
>
> 4. Reproducibility (inter-architecture)

The combination of these trustworthiness criteria provide a blueprint for us to assess a saliency map's localization capabilities (localization utility), sensitivity to trained model weights (versus randomized weights), and robustness with respect to models trained with the same architectures (repeatability) and different architectures (reproducibility). Fig 1 summarizes the questions that will be addressed in this work.

## Materials and Methods

### Data Preparation

The Pneumothorax dataset consists of 10675 images, split in a 81 : 9 : 10 ratio to create training, validation and test sets respectively. The training set is comprised of 8646 images with 1931 positive cases; the validation set contains 961 images with 202 positive cases and the test set contains 1068 images with 246 positive cases.

The Pneumonia dataset consists of 14863 images, which is split in a similar fashion as described above for the Pneumothorax dataset. The final training set is comprised of 12039 images with 4870 positive cases; the validation set contains 1338 images with 541 positive cases and the test set contains 1486 images with 601 positive cases.

### Model Training

We train InceptionV3 [38] and Densenet121 [39] models for all of our experiments. The InceptionV3 network is comprised of several modules, allowing for more efficient computation and deeper networks through dimensionality reduction with stacked 1×1 convolutions. The Densenet121 network is comprised of four blocks with 6, 12, 24 and 16 layers each, which extract features to be sent to a classification module. These image classification model backbones were modified by replacing the final layer to perform binary classification. The models were loaded with pretrained weights on ImageNet, which were then finetuned during training. Both InceptionV3 and DenseNet121 models are trained using Binary Cross Entropy loss:

$$H_p(q) = -\frac{1}{N}\sum_{i=1}^{N}(y_i \log p(y_i) + (1-y_i)\log(1-p(y_i)))$$

Further details about model training are deferred to the supplement.

### Saliency Methods and Evaluation Criteria

For model interpretability, we evaluate the following saliency methods: Gradient Explanation (GRAD), Smoothgrad (SG), Integrated Gradients (IG), Smooth IG (SIG), GradCAM, XRAI, Guided-backprop (GBP), and Guided GradCAM (GGCAM). All methods are summarized and defined in Table 1. We compared the performance of these saliency maps against the following baselines: a) for localization utility, a low baseline defined by a single "average" mask of all ground truth segmentations/bounding boxes in the training and validation datasets, and a high baseline determined by the Area Under the Precision Recall Curve (AUPRC [20])[1] of segmentation (U-Net) and detection networks (RetinaNet); b) in model weight randomization, the average Structural SIMilarity (SSIM)[2] indices of 50 randomly chosen pairs of saliency maps pertaining to the fully trained model; and c) in repeatability and reproducibility, a low baseline of SSIM=0.5, and a high baseline determined by the SSIM of U-Net and RetinaNet. Note that a low baseline of SSIM=0.5 is chosen because SSIM ranges from 0 (for lack of any structural similarity) to 1 (for identical structural similarity), and SSIM=0.5 marks the midpoint for whether the SSIM is more structurally similar or dissimilar[32].

---

[1] Precision recall curves better serve to be more informative about an algorithm's performance, especially for unbalanced datasets with few positive pixels relative to the number of negative pixels. In the context of findings on medical images, the area under the Receiver Operator Characteristic (ROC) curve can be skewed by the presence of a large number of true negatives [25]

[2] SSIM is a metric used for evaluating image similarity computed as a weighted combination of the comparison measurements of luminance, contrast and structure. SSIM=1 is achieved when comparing identical sets of data whereas SSIM=0 indicates no structural similarity.

Although saliency maps are not originally intended for either segmentation or detection, they are being used this way in clinical research to identify areas of abnormality from trained neural networks [3, 4, 5]. Using saliency maps in this manner can cause potential problems when this type of research is applied to clinical practice. Therefore, we chose to evaluate saliency maps using pixel-based metrics to show the discrepancies with using them in such a manner, and provide objective measures as to why using saliency maps in place of segmentation or detection is a dangerous practice. We choose AUPRC as the metric to capture localizability of saliency maps as the relatively small size of the ground truth segmentation masks and bounding boxes necessitated an approach that would account for the class imbalance.

Precision is defined as the ratio of True Positives (TP) to Predicted Positives (TP+FP) while Recall is defined as the ratio of True Positives to Ground Truth Positives (TP+FN). Since neither of these account for the number of True Negatives, they make ideal candidates for our analysis. To capture the intersection between the saliency maps and segmentation masks or bounding boxes, we consider the pixels inside the segmentation/boxes to be positive labels and those outside to be negative. Each pixel of the saliency map is thus treated as an output from a binary classifier. An ideal saliency map, from the perspective of utility would have perfect recall (finding all regions of interest) without labelling any pixels outside the regions of interest as positive (perfect precision).

To investigate the sensitivity of saliency methods under changes to model parameters and identify potential correlation of particular layers to changes in the maps, we employ cascading randomization on the InceptionV3 model[3]. In cascading randomization, we successively randomize the weights of the model beginning from the top layer to the bottom, effectively erasing the learned weights in a gradual fashion. We use the Structural SIMilarity (SSIM) index to assess the change of the original saliency map with the saliency maps generated from the model after each randomization step[21].

Training details of the models, corresponding baselines (U-Net for segmentation, RetinaNet for detection), and additional description of the utility metric (AUPRC) are provided in the Supplemental Material.

**Statistical analysis**

Statistical analyses were performed in RStudio version 1.2.5033 using R 3.6 and the *lmer* (lme4, v.1-1-25), *lmerTest* (3.1-3), *ggplot2* (3.3.2) and *multComp* (1.4-14) packages. A linear mixed effects model was used to evaluate trustworthiness of the eight saliency map methods using a two-sided test with alpha level set at 0.05 for statistical significance. Tukey's HSD test was used for post-hoc analysis. For our statistical analysis, the following hypothesis questions were examined:

1. Is there a difference between the utility of each of the saliency map methods derived from the trained classification models compared to the localization baseline models (detection or segmentation) as measured using AUPRC?

2. Is there a difference between the trained models and the "average" mask (i.e., average of training and validation segmentations or bounding boxes) in terms of the utility of the map as measured using AUPRC?

3. Is there a difference in the output of trained models compared to their associated random model as measured using SSIM?

4. Is there a difference between the repeatability/reproducibility of each of the saliency map methods compared to the localization baselines as measured using SSIM?

# Results

## Localization Utility

### Segmentation Utility

We evaluate the localization utility of each saliency method by quantifying their intersection with ground truth pixel-level segmentations available from the Pneumothorax dataset. We compare the saliency methods with the average of the segmentations across the training and validation sets, as well as a vanilla U-Net[4] trained to learn these segmentations directly.

Saliency maps generated from InceptionV3 demonstrate better test set results than those generated from DenseNet121, and are displayed in Fig 2a. For individual saliency maps on InceptionV3, the best performing method is XRAI (AUPRC=$0.224 \pm 0.240$), while the worst performing method is SIG (AUPRC=$0.024 \pm 0.021$). It is also interesting to note that using the average of all masks across the pneumothorax training and validation datasets (AVG) performs as well or better than most of the saliency methods (AUPRC=$0.142 \pm 0.149$), showing a strong limitation in the saliency maps' utility.

Specifically, GBP, GCAM, and GGCAM are not statistically different than the average map, GRAD ($p=0.0279$), IG ($p<0.005$), SG ($p<0.005$), SIG ($p<0.005$) are all significantly worse and XRAI is significantly better ($p<0.005$).

Additionally, the U-Net trained on a segmentation task achieves the best performance by far (AUPRC=$0.404 \pm 0.195$) and the utility of all maps are significantly worse than the U-Net ($p<0.005$).

The utility of the saliency maps generated using the trained models was higher than the utility of the random models ($p<0.005$) in all cases except for SG and SIG, where there were no statistical differences.

### Detection Utility

We evaluate detection utility of each saliency method using the ground truth bounding boxes from the Pneumonia Detection dataset. Fig S1, in the Supplemental Material, shows visualizations of saliency maps generated from InceptionV3 on the Pneumonia Detection dataset. In the same fashion as the Segmentation Utility subsection, we calculate the AUPRC considering pixels inside the bounding boxes to be positive labels and those outside to be negative. We compare the saliency methods with the average of the bounding boxes across the training and validation sets, as well as a RetinaNet[30] trained to learn these bounding boxes directly.

Results for the test set are shown in Fig 2b. The best performing saliency method is XRAI (AUPRC=$0.519 \pm 0.220$), while the worst performing method is SIG (AUPRC=$0.160 \pm 0.128$). It is interesting to note that using the average of all bounding boxes across the Pneumonia training and validation datasets performs better than all the methods (AUPRC=$0.465 \pm 0.268$) except for XRAI, which was significantly better ($p<0.005$). RetinaNet trained to generate bounding boxes achieves significantly better performance than all the saliency methods (AUPRC=$0.596 \pm 0.260, p<0.005$).

The utility of saliency maps generated using the trained models was higher than the utility of the random models for GRAD, GCAM, XRAI and GGCAM ($p<0.005$). The random model had higher utility for SG and SIG ($p<0.005$) and there were no statistical differences for IG ($p=0.83$) and GBP ($p=0.6$).

## Sensitivity to Trained vs Random Model Weights

Saliency maps should be sensitive to model weights in order to be meaningful. Specifically, a saliency map generated from a trained model should differ from a randomly initialized model, which has no knowledge of the task.

Fig 3a shows the progressive degradation of saliency maps and Fig 3b shows an example image of saliency map degradation from cascading randomization. Fig S2, in the Supplemental Material, shows additional examples. Table S1, in the Supplemental Material, shows the mean SSIM index scores of saliency maps for fully randomized models (after cascading randomization has reached the bottom-most layer), as well as the corresponding degradation thresholds, defined below.

A saliency map has reached degradation and is classified as a 'PASS' when the average SSIM goes below the degradation threshold, which we define as the average SSIM of 50 pairs of randomly chosen saliency maps for a given saliency method on the test dataset. This baseline was chosen because the SSIM for the saliency map should be at least as different as the one produced for a completely different input image using the same saliency method. For both pneumothorax and pneumonia datasets, we observe that the saliency maps that fall below this degradation threshold when cascading randomization has reached the bottom-most layer (i.e., fully randomized) include GradCAM, GBP, and GGCAM, showing a dependency on the model weights, which is desired. We also note that although XRAI performed the best in the localization tests, for both pneumothorax and pneumonia datsets, XRAI does not reach the degradation threshold when fully randomized, showing invariance to the trained model parameters.

**Repeatability and Reproducibility**

We conduct repeatability tests on the saliency methods by comparing maps from a) different randomly initialized instances of models with the same architecture trained to convergence (intra-architecture repeatability) and b) models with different architectures each trained to convergence (inter-architecture reproducibility) using SSIM between saliency maps produced from each model. These experiments are designed to test if the saliency methods produce similar maps with a different set of trained weights and whether they are architecture agnostic (assuming that models with different trained weights or architectures have similar classification performance). Although there is no constraint that indicates that interpretations should be the same across models, an ideal trait of a saliency map would be to have some degree of robustness across models with different trained weights or architectures. For comparison, we have a low baseline of SSIM=0.5 (since SSIM=0.5 marks the midpoint for whether the SSIM is more structurally similar or dissimilar), and a high baseline of repeatability/reproducibility of separately trained U-Nets (for the pneumothorax dataset) and RetinaNets (for the pneumonia dataset).

We examine the repeatability of saliency methods from two separately trained InceptionV3 models, and the reproducibility of saliency methods from trained InceptionV3 and DenseNet121 models. Figs 4a and 4b summarize the results for the Pneumothorax and Pneumonia datasets, respectively. In the Pneumothorax dataset, the baseline U-Net achieves a value of SSIM=$0.976 \pm 0.024$, which is much greater performance than any of the saliency maps ($p<0.005$). The lower SSIM=0.5 baseline is greater than the performance of all saliency maps except for the repeatability of XRAI and GGCAM. Among the saliency maps, XRAI has the best repeatability (SSIM=$0.643 \pm 0.092$), while SG has the worst repeatability (SSIM=$0.176 \pm 0.027$). For reproducibility, XRAI performs the best (SSIM=$0.487 \pm 0.084$), while GRAD performs the worst (SSIM=$0.169 \pm 0.013$). Repeatability was higher than reproducibility for GRAD, IG, XRAI, GBP, and GGCAM. Surprisingly, reproducibility was higher than repeatability for SG and SIG.

In the Pneumonia dataset, the baseline RetinaNet achieves a value of SSIM=$0.802 \pm 0.048$, which is only exceeded by XRAI's repeatability score ($p<0.005$). The lower SSIM=0.5 baseline is greater than the performance of all saliency maps except the repeatability and reproducibility of GCAM, GGCAM, and XRAI, and the repeatability of GBP. Among the saliency maps, XRAI has the best repeatability (SSIM=$0.836 \pm 0.055$), while SG has the worst (SSIM=$0.270 \pm 0.009$). For reproducibility, XRAI performs the best (SSIM=$0.754 \pm 0.062$), while SG performs the worst (SSIM=$0.184 \pm 0.014$). Repeatability was higher than reproducibility for all methods ($p \ll 0.005$). Overall, XRAI has the highest repeatability and reproducibility across different datasets. In Figs 4c and 4d, we observe the repeatability and reproducibility of two example images from the Pneumothorax and Pneumonia datasets, respectively.

The first two rows are saliency maps generated from two separately trained InceptionV3 models (Replicates 1 and 2) to demonstrate repeatability, and the last row are saliency maps generated DenseNet121 to demonstrate reproducibility. Fig S3, in the Supplemental Material, shows additional examples for repeatability and reproducibility.

## Discussion and Conclusion

In this study, we evaluated the performance and robustness of several popular saliency maps on medical images. By considering utility with respect to localization, sensitivity to model weight randomization, repeatability and reproducibility, we demonstrate that none of the saliency maps meet all tested criteria and their credibility should be critically evaluated prior to integration into medical imaging pipelines. This is especially important because many recent deep learning based clinical studies rely on saliency maps for interpretability of deep learning models without noting and critically evaluating their inherent limitations. A recent empirical study found that ophthalmologists and optometrists rated GBP highly as an explainability method, despite the limitations we note in this study[33]. Table 2 demonstrates the overall results for each saliency map across all tests. A 'PASS' is denoted if the saliency method performs better than the respective baseline, while a 'FAIL' is denoted if the saliency method performs worse than the respective baseline (described further in Table 2). It is clear that none of the maps demonstrate a superior performance in all four defined trustworthiness criteria, and in fact most of them are inferior to their corresponding baselines. For their high baseline methods, the utility, repeatability, and reproducibility tasks utilize networks that train specifically as localizers (i.e., U-Net and RetinaNet). With the exception of XRAI on repeatability in the pneumonia dataset, all the saliency maps perform worse than U-Net and RetinaNet. This highlights a severe limitation in the saliency maps as a whole, and shows that using models trained directly on localization tasks (such as U-Net and RetinaNet) greatly improves the results. Additional insights for the results of each task are provided in the Supplemental Material.

Depending on the desired outcome of interpretability, there are alternative techniques besides saliency methods that can be employed. One approach would be to train CNNs that output traditional handcrafted features (such as shape and texture) as intermediates[28]. This approach would provide some interpretability but is limited by the utility and reliability of the handcrafted features. Another approach may also be to use interpretable models in the first place. Rudin argues that instead of creating methods to interpret black box models trained for high-stakes decision making, we should instead put our focus on designing models that are inherently interpretable[23]. Rudin further argues that there is not necessarily a tradeoff between accuracy and interpretability, especially if the input data is well structured (i.e., features are meaningful). Thus moving forward from the results in this work, it would be wise to consider multiple avenues for improving model interpretation.

There are a few limitations to our study. First, we only evaluated saliency maps for two medical datasets, both consisting of chest radiographs. Future studies will examine more medical imaging datasets, including different image modalities and diseases. Additionally, we only performed tests on two CNN architectures, though these are commonly used networks in the literature for chest radiograph analysis[13, 14]. As a next step, we can examine the effect of other CNN architectures to determine if they result in saliency maps that are more repeatable and reproducible. Third, we focus only on the ability of saliency maps to localize pathology and thus the utility metrics were calculated using the regions-of-interest specifically (bounding boxes for pneumonia and segmentation maps for pneumothorax). These regions-of-interest may not include other image features that can contribute to classification algorithm performance, known as hidden stratification. For example, a chest tube in an image would imply the presence of a pneumothorax, but much of the chest tube may not be in the region-of-interest[31]. More global features could also contribute to classification. For example, low lung volumes and portable radiograph technique may suggest that the patient is hospitalized, which could be associated with likelihood of pneumonia. These features would also not be covered in the regions-of-interest. Future work can evaluate the utility of saliency maps to localize these other features. We could also investigate incorporating saliency maps as a part of neural network training and evaluate if this type of approach results in maps that have higher utility than maps that are generated post-model training[24].

## Competing Interests

J.K.C. has research funding from General Electric. All other authors declare that there are no competing interests.

## Author Contribution

N.A., N.G., P.S., and K.C. wrote the main manuscript. N.A., N.G., P.S., K.C., and J.K.C. led the development of the testing criteria of saliency maps. M.A. developed the Retina-Net benchmark. B.C. developed the U-Net benchmark. J.A. and N.A. developed the cascading randomization module. M.D.L. ensured medical appropriateness of segmentations and bounding boxes. N.A., N.G., P.S., K.C., M.A., B.C., K.H., S.G., J.P., M.G., J.A., M.D.L., and J.K.C. participated in the conceptualization of the saliency map testing criteria. All authors reviewed and approved this manuscript.

## Acknowledgement

The authors acknowledge the importance of the work of authors P.S. and K.C. in the preparation of this article. They were instrumental in the conception and implementation of this work.

## Data Availability

The datasets in this study are publicly available and links can be found in the references [1, 2, 3, 4]. The code used for all tests is also available at https://github.com/QTIM-Lab/Assessing-Saliency-Maps.

# Tables and Figures

## Tables

**Table 1:** Saliency methods evaluated in this work, along with their corresponding definitions

| Saliency Map | Definition |
|---|---|
| **Gradient Explanation (GRAD)** [15] | Measures the extent to which a change in a region of the input *x* affects the prediction $S(x)$ to compute the map $\frac{\partial S}{\partial x}$. |
| **Smoothgrad (SG)** [18] | Smoothes the mask obtained using the gradient and integrated gradient saliency methods by stochastically modifying input and performing Gaussian smoothing on the resulting maps. |
| **Integrated Gradients (IG)** [16] | Constructs a map by interpolating from a baseline image to the input image and averaging the gradients across these interpolations. We use 25 such interpolations in our experiments to compute the masks. |
| **Smooth IG (SIG)** [16, 18] | Smoothes an integrated gradients map by stochastically modifying input and performing Gaussian smoothing. |
| **GradCAM (GCAM)** [7] | A backpropagation-based method that uses the feature maps of the final convolutional layer to generate heatmaps. |
| **XRAI** [6] | Builds on integrated gradients by starting with a baseline image and incrementally adding regions that offer maximal attribution gain. |
| **Guided-backprop (GBP)** [19] | Constructs a mask obtained by 'guiding' the conventional backpropagation algorithm to suppress any negative gradients. |
| **Guided GradCAM (GGCAM)** [7] | Combines the masks obtained by GradCAM and Guided-backprop in an attempt to minimize the false positive produced by either. |

**Table 2:** Summary of all the results for experiments on the (a) SIIM-ACR Pneumothorax segmentation dataset and (b) RSNA Pneumonia detection dataset. A 'PASS' is denoted if the saliency method performs better than the respective baseline, while a 'FAIL' is denoted if the saliency method performs worse than the respective baseline (described below). The Utility column compares maps to the average of all maps across the training and validation datasets (AVG) and a U-Net (UNET) trained for pneumothorax segmentation, or a RetinaNet (RNET) trained for pneumonia detection. The Randomization column compares the SSIM scores of the saliency maps when cascading randomization is completed to the bottom-most layer with the degradation threshold defined in the Results section. The Repeatability and Reproducibility columns compare SSIM scores of saliency maps with the low baseline of SSIM = 0.5 (LOW) and two independently trained U-Nets for pneumothorax segmentation or two independently trained RetinaNets for pneumonia detection.

| Method | Utility | | Randomization | Repeatability | | Reproducibility | |
|---|---|---|---|---|---|---|---|
| | AVG | UNET | | LOW | UNET | LOW | UNET |
| GRAD | FAIL | FAIL | FAIL | FAIL | FAIL | FAIL | FAIL |
| SG | FAIL | FAIL | FAIL | FAIL | FAIL | FAIL | FAIL |
| IG | FAIL | FAIL | FAIL | FAIL | FAIL | FAIL | FAIL |
| SIG | FAIL | FAIL | FAIL | FAIL | FAIL | FAIL | FAIL |
| GCAM | FAIL | FAIL | PASS | FAIL | FAIL | FAIL | FAIL |
| XRAI | PASS | FAIL | FAIL | PASS | FAIL | FAIL | FAIL |
| GBP | FAIL | FAIL | PASS | FAIL | FAIL | FAIL | FAIL |
| GGCAM | FAIL | FAIL | PASS | PASS | FAIL | FAIL | FAIL |

(a) SIIM-ACR Pneumothorax Segmentation

| Method | Utility | | Randomization | Repeatability | | Reproducibility | |
|---|---|---|---|---|---|---|---|
| | AVG | UNET | | LOW | Method | AVG | UNET |
| GRAD | FAIL | FAIL | PASS | FAIL | FAIL | FAIL | FAIL |
| SG | FAIL | FAIL | PASS | FAIL | FAIL | FAIL | FAIL |
| IG | FAIL | FAIL | FAIL | FAIL | FAIL | FAIL | FAIL |
| SIG | FAIL | FAIL | PASS | FAIL | FAIL | FAIL | FAIL |
| GCAM | FAIL | FAIL | PASS | PASS | FAIL | PASS | FAIL |
| XRAI | PASS | FAIL | FAIL | PASS | PASS | PASS | FAIL |
| GBP | FAIL | FAIL | PASS | PASS | FAIL | FAIL | FAIL |
| GGCAM | FAIL | FAIL | PASS | PASS | FAIL | PASS | FAIL |

(b) RSNA Pneumonia Detection

**Figures**

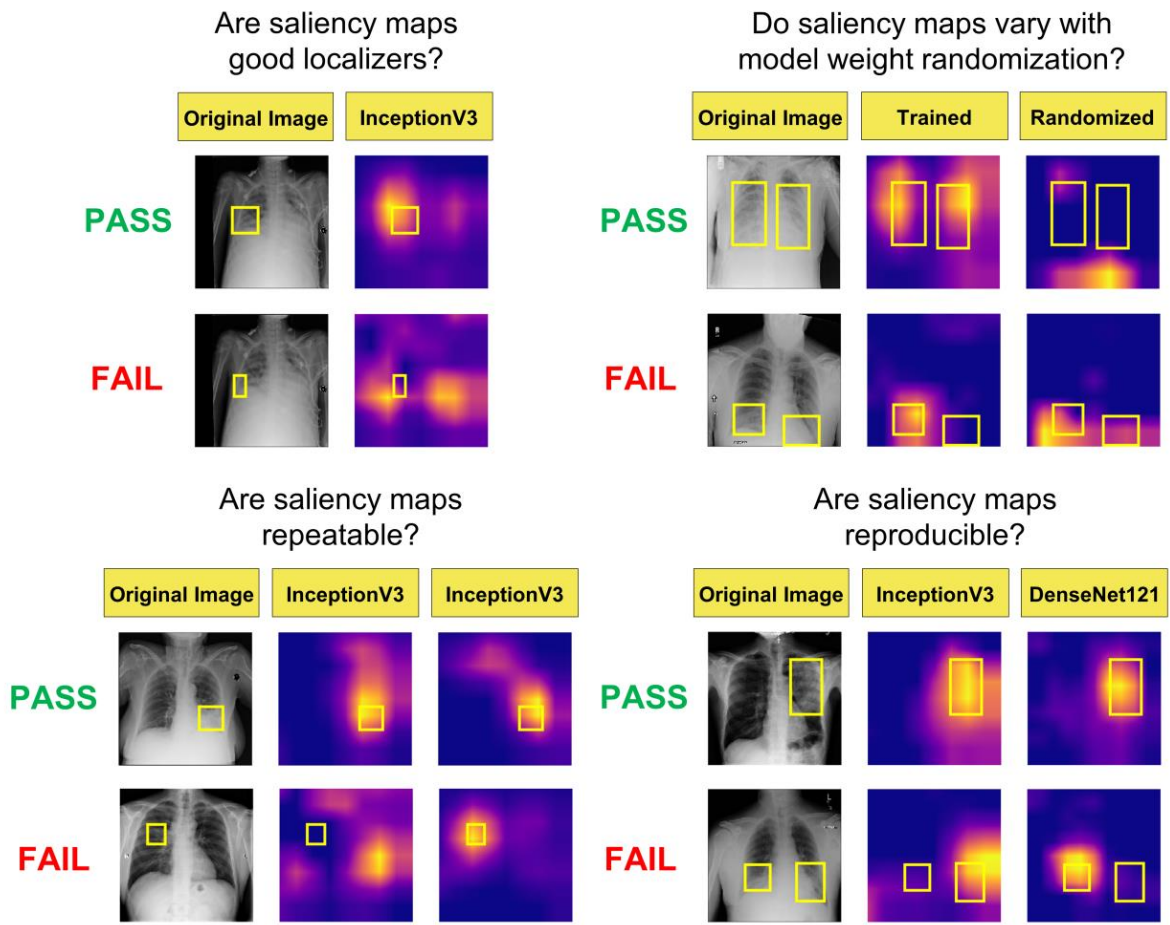

**Fig. 1:** Visualization of the different questions that will be addressed in this work. Note that the top rows of images and saliency maps demonstrate ideal (and less commonly observed) high performing examples ('PASS'), while the bottom rows of images demonstrate realistic (and more commonly observed) poor performing examples ('FAIL'). First, we examine whether saliency maps are good localizers (in regards to the extent of the maps' overlap with pixel-level segmentations or ground truth bounding boxes). Next, we evaluate whether saliency maps are affected when trained model weights are randomized, indicating how closely the maps reflect model training. Then we generate saliency maps from separately trained InceptionV3 models to assess their repeatability. Finally, we assess the reproducibility by calculating the similarity of saliency maps generated from different models (InceptionV3 and DenseNet121) trained on the same data.

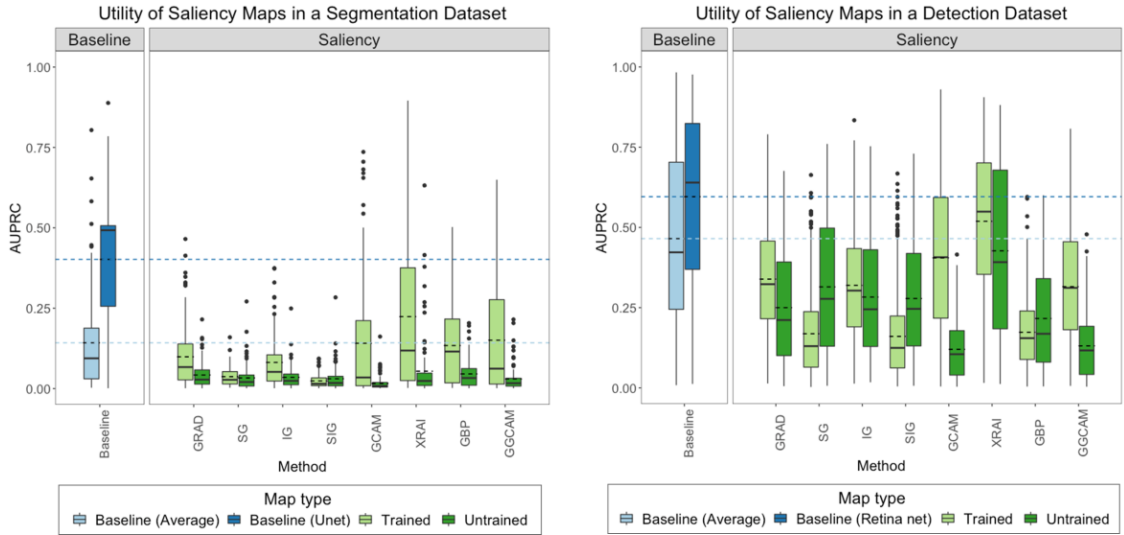

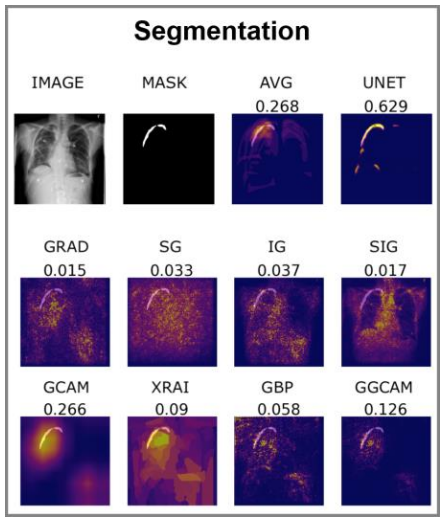        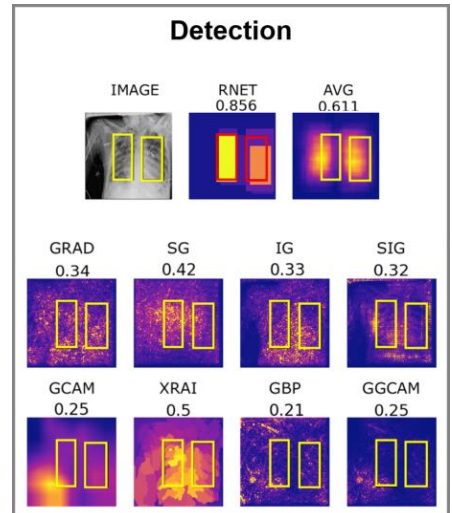

(c)                                    (d)

**Fig. 2:** (a) test set segmentation utility scores for SIIM-ACR Pneumothorax segmentation dataset; (b) test set bounding box detection utility scores for RSNA Pneumonia detection dataset; each box plot represents the distribution of scores across the test data sets for each saliency map, with a solid line denoting the median and dashed line denoting the mean; results are compared to a low baseline using the average segmentation/bounding box of the training and validation sets (light blue) and high baseline using U-Net/RetinaNet (dark blue); (c) example saliency maps on SIIM-ACR Pneumothorax dataset with corresponding utility scores; (d) example saliency maps on RSNA Pneumonia dataset with corresponding utility scores; "Average Mask"/"AVG" refers to using the average of all ground-truth masks (for pneumothorax) or bounding boxes (for pneumonia) across the training and validation datasets; "U-Net"/"UNET" refers to using the vanilla U-Net trained on a segmentation task for localization of pneumothorax; "RetinaNet"/"RNET" refers to using RetinaNet to generate bounding boxes for localizing pneumonia with bounding boxes.

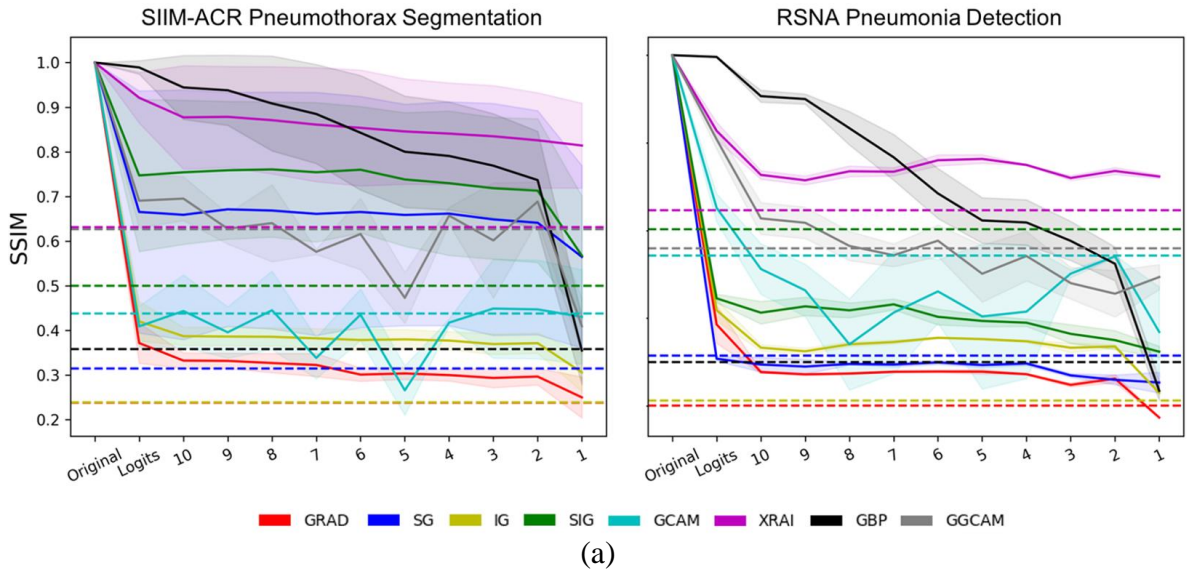

(a)

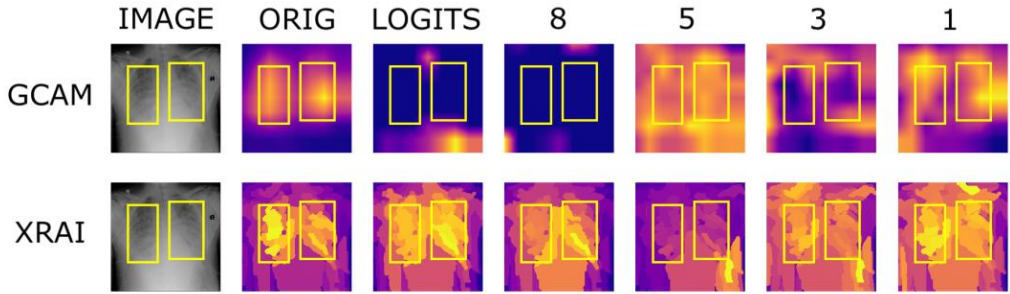

(b)

**Fig. 3:** (a) Structural SIMilarity (SSIM) index under cascading randomization of modules on InceptionV3 for SIIM-ACR Pneumothorax segmentation dataset and RSNA Pneumonia detection dataset; note that the colored dotted lines correspond to the degradation threshold for each saliency map; they are generated by the average SSIMs of 50 randomly chosen pairs of saliency maps pertaining to the fully trained model; a saliency model successfully reaches degradation if it goes below its corresponding degradation threshold; (b) example image from RSNA Pneumonia detection dataset to visualize saliency map degradation from cascading randomization; 'Logits' refers to the Logit Layer (final layer) of the InceptionV3 model and layer blocks 1 through 10 refer to blocks Mixed1 through Mixed10 in the original InceptionV3 architecture.

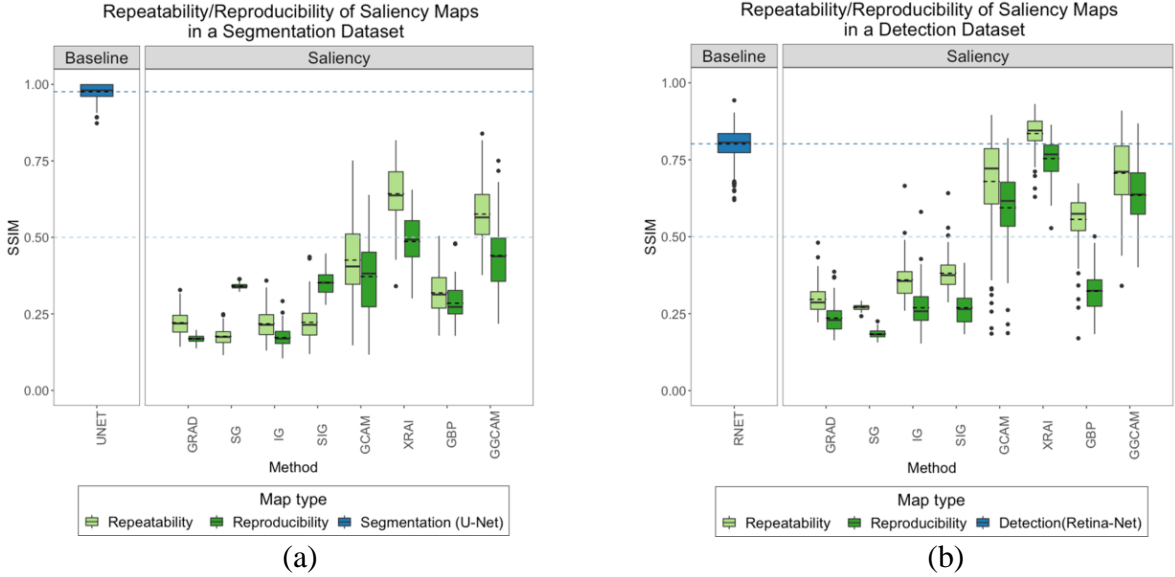

(a)           (b)

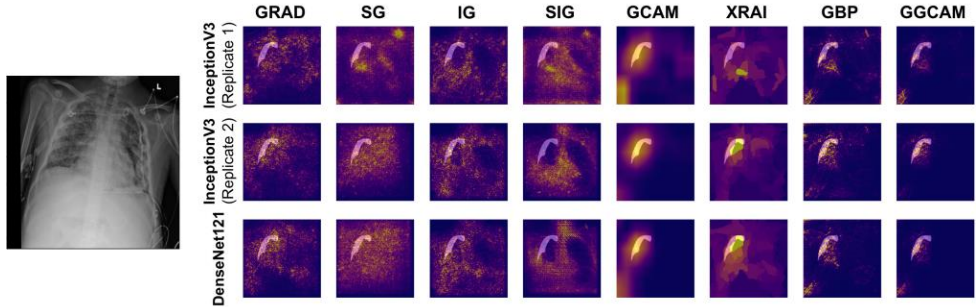

(c)

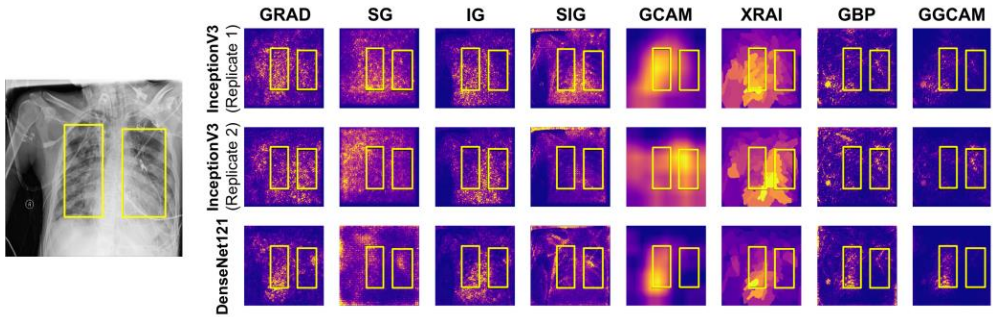

(d)

**Fig. 4:** Comparison of repeatability and reproducibility scores for all saliency methods for (a) SIIM-ACR Pneumothorax segmentation dataset and (b) RSNA Pneumonia detection dataset; each box plot represents the distribution of scores across the test data sets for each saliency map, with a solid line denoting the median and dashed line denoting the mean; results are compared to a low baseline of SSIM = 0.5 (light blue dashed line) and high baseline using U-Net/RetinaNet (dark blue box plot and dashed line); two examples of repeatability (InceptionV3 Replicates 1 and 2) and reproducibility (InceptionV3 and DenseNet121) for the (c) SIIM Pneumothorax dataset with transparent segmentations and (d) RSNA Pneumonia dataset with yellow bounding boxes

# Supplemental Material

## Abbreviations

- **2D**: Two-dimensional
- **ACR**: American College of Radiology
- **AUC**: Area Under the Curve
- **AUPRC**: Area Under the Precision-Recall Curve
- **AUROC**: Area Under the Receiver Operating Characteristic Curve
- **AVG**: Average of all masks (bounding boxes/segmentations) across the training and validation datasets
- **CNN**: Convolutional Neural Network
- **FN**: False Negatives
- **FP**: False Positives
- **GBP**: Guided-backprop
- **GGCAM**: Guided Gradient-weighted Class Activation Mapping
- **GRAD**: Gradient Explanation
- **GradCAM**: Gradient-weighted Class Activation Mapping
- **IG**: Integrated Gradients
- **LOW**: Low baseline
- **PR**: Precison-Recall
- **ReLU**: Rectified Linear Unit
- **RNET**: RetinaNet
- **ROC**: Receiver Operator Characteristic
- **RSNA**: Radiological Society of North America
- **SG**: Smoothgrad
- **SIG**: Smooth IG
- **SIIM**: Society for Imaging Informatics in Medicine
- **SSIM**: Structural Similarity Index Measure
- **TP**: True Positives
- **UNET**: U-Net

## Supplemental Methods and Materials

### Model Training

We train the InceptionV3 models used for our analysis via transfer learning using ImageNet-pretrained model weights [1] by replacing the final fully connected layer (https://github.com/keras-team/keras-applications/blob/master/keras_applications/inception_v3.py). The learning rate is set at 1*e*- 4 after experimenting with different values, and the models are trained for a maximum of 20 epochs. Early stopping [2] is used while monitoring the validation AUC with a patience of 4 epochs. We use the TensorFlow framework for training all of our models.

In a similar fashion, we use an ImageNet-pretrained DenseNet121 model (https://github.com/keras-team/keras-applications/blob/master/keras_applications/densenet.py) for reproducibility analysis. The learning rate is set at 7*e*- 5 and the models are trained for a maximum of 30 epochs. Early stopping [2] is used while monitoring the validation AUC with a patience of 5 epochs. Both the InceptionV3 and the DenseNet121 models are trained using Binary Cross Entropy loss as given below

$$H_p(q) = -\frac{1}{N} \sum_{i=1}^{N} (y_i \log p(y_i) + (1 - y_i) \log(1 - p(y_i)))$$

Cascading randomization is performed on the InceptionV3 model to reset the parameter values sampled from a truncated normal distribution, in accordance with Adebayo et al. [3]. The models trained on the SIIM-ACR Pneumothorax dataset report test set AUCs of 0.889 (InceptionV3) and 0.909 (DenseNet121), while the fully trained models on the curated RSNA Pneumonia dataset report test set AUCs of 0.936 (InceptionV3) and 0.927 (DenseNet121). The corresponding completely random models both record AUCs of 0.5 as expected. The trained InceptionV3 replicate used for the repeatability experiment on the RSNA Pneumonia dataset reported a test set AUC of 0.940 and the correspoding replicate for the SIIM-ACR Pneumothorax dataset reported a test set AUC of 0.907.

For the segmentation baseline model, we use the base U-Net architecture [4] with no pretraining. During training, the Adam optimizer is used with default beta values, a mix of focal and dice loss, batch size of 4, and learning rate of 1e-04. The learning rate is decreased by a factor of 0.1 when the validation loss fails to decrease for more than 3 epochs. The best model (based on the validation loss) is trained for up to 75 epochs, using early stopping if the loss has not decreased for 15 epochs. Additionally, in order to equally sample the positive and negative pneumothorax cases, balanced sampling is used on the images. For the final output, the sigmoid function is applied to each cell (pixel). The mean non-zero cell value in this output is taken (if it exists, otherwise we output 0) and used as the model's classification output for calculating the AUC of the model. We obtain a test AUC score of 0.893. To generate the final image, the sigmoided U-Net output is multiplied by 255 and rounded to scale the values into a final output image.

For the object detection baseline model, we use RetinaNet with ResNet101 base architecture pretrained on ImageNet (https://github.com/yhenon/pytorch-retinanet.git). During training, the Adam opimizer is used with hyper-parameters set at 1e-4 learning rate, 4 batch-size and, 0.9 and 0.999 beta values. To get a pixel-wise output for evaluation, a continuous mask is constructed with all detection boxes having greater than 0.01 probability. Similar to how the non-maximum suppression algorithm works for appropriate box selection, each pixel is assigned a probability value corresponding to the highest scoring box covering that pixel. The result is a heatmap where each pixel value is directly associated with probability of presence of abnormality in that pixel region. We obtain a test AUROC of 0.949 and test AUPRC of 0.932.

**Utility Metric**

We choose the area under the precision recall curve (AUPRC) as the metric to capture localizability of saliency maps as the relatively small size of the ground truth segmentation masks and bounding boxes necessitated an approach that would account for the class imbalance.

A PR curve shows the trade-off between precision and recall across different decision thresholds by varying the threshold to plot corresponding precision and recall values. An example curve for a saliency map is as shown below in Fig S4 in the Supplementary Material.

**Supplemental Discussion**

In regards to evaluating localization utility for segmentation of pneumothorax images, maximum utility is achieved by XRAI at AUPRC = $0.224 \pm 0.240$. Additionally, in case of evaluating localization utility for detection of pneumonia images, maximum utility is achieved by XRAI at AUPRC = $0.519 \pm 0.220$. Although this suggests a substantial improvement from the pneumothorax results, it is important to observe that ground truth bounding boxes for detection tasks do not have complex shapes and cover a superset of pixels from a corresponding mask for segmentation tasks. This leads to less difficult localization task.

In general, the particular localization tasks presented in this paper can be incredibly difficult due to the overlapping structures present in 2D chest radiographs, as well as subtle changes in texture that can be challenging to detect [5]. The challenges of the pneumothorax and pneumonia chest radiograph datasets serve to demonstrate the limitations on localization abilities of saliency maps. These findings highlight the necessity to build saliency maps that better consider complex shapes of the object of interest (segmentation utility), as well as have improved localization capabilities in general (detection utility). To inform future saliency map development, we can consider some aspects of the better performing ones. Among the saliency maps that were tested, XRAI performed the best for both types of utility tasks. For each test image in the dataset, XRAI segments the image into small regions, iteratively evaluates the relevance of each region to the model prediction, and aggregates the smaller regions into a larger region based on the relevance scores [6]. This iterative evaluation of small patches within the image likely gives XRAI an advantage over other methods, since it results in maps with better fine-grained localization catering to adjacent spatial neighborhoods, thus achieving a higher recall and precision than the other methods. In the detection utility task, GradCAM performed the second best. GradCAM [7] generates maps with smooth activations (since activations are derived from gradients that are up-sampled from a lower-dimensional convolutional layer). This smoothness likely plays a role in having an improved AUPRC since the map is less granular compared to other saliency maps. GradCAM's smoothness property provides a strong advantage with high recall since more activations within the area covered by the segmentations or bounding boxes likely have higher values. However, with segmentations being more granular than the up-sampled GradCAM map, there would be a more severe penalty when GradCAM shows high activations in areas outside the segmented area. Therefore, GradCAM would be safer to use on applications with less granular localization, such as a detection task (with bounding boxes). When considering how to build a saliency map with improved pixel-level localization, some of these aspects of XRAI and GradCAM may be of value to consider.

Additionally, for both datasets, under the application of cascading randomization across the different layers of the InceptionV3 model, GradCAM and GGCAM demonstrate a degraded SSIM that goes below the previously defined degradation threshold. This demonstrates GradCAM's sensitivity to trained model weights versus randomly initialized weights, which should be an important consideration when choosing a saliency map to use for prediction interpretation (if a saliency map does not change considerably after trained model weights have been randomized, this implies that the saliency map is not model weight-sensitive). GradCAM forward propagates test images through the model to obtain a prediction, then backpropagates the gradient of the predicted class to the the desired convolutional feature map [7]. As a result, there is a high sensitivity to the value of the weights in the model. In contrast, other saliency methods that do not show as much sensitivity to the value of model weights are more affected by details in the image input (e.g., contrast, sharpness, etc.) than by the trained model weights. Although GBP demonstrated degradation when cascading randomization reached the bottom-most layer, the SSIM remained above the degradation threshold across all other layers indicating a limitation of GBP's sensitivity to trained weights. There is a study that found that GBP performs a partial image recovery that is unrelated to the trained neural network's decisions [8]. The study found that backward ReLU and local connections in the trained CNNs are two main contributors to this effect. In regards to local connections, with the exception of neurons in the first layer of a CNN, each neuron is only associated with a small, local group of pixels from the input image (i.e., all the input that goes into a neuron is spatially close to each other). When backward ReLU is performed, theoretical analysis has shown that local connections cause the resulting visualizations to be more human-interpretable but less class-sensitive [8]. Properties like these should be accounted for when developing a saliency technique. Note that we did not take the absolute values of the maps before computing the SSIM and instead used the raw saliency maps to get a clearer picture of map stability.

Finally, for both datasets, XRAI demonstrates the highest repeatability score between two separately trained models with the same architecture and also the highest reproducibility between two separately trained models with different architectures. XRAI's aggregation of smaller regions into larger regions likely reduces the influence of variability across trained models with similar architectures. Thus, the overall model weight distribution should remain the same in a specific area for a particular image even if the models are separately trained. The aggregation of smaller regions to larger regions likely also has a stabilizing effect even on models of different architectures. However, these properties have not yet been extensively studied since XRAI is a fairly new saliency method. It is also notable that the spatial resolution of the generated heatmaps are not the same across methods. In particular, GradCAM's resolution depends on the dimensions of the final convolutional layer of the network, which is up-sampled to fit the dimensions of the model input images (in our case the final dimensions of the DenseNet121 and InceptionV3 models is $10 \times 10$) [7]. A single activation change in the $10 \times 10$ feature map might result in a high degree of variability in the corresponding enlarged GradCAM generated map, hence influencing the repeatability and reproducibility scores.

**Supplemental Figures**

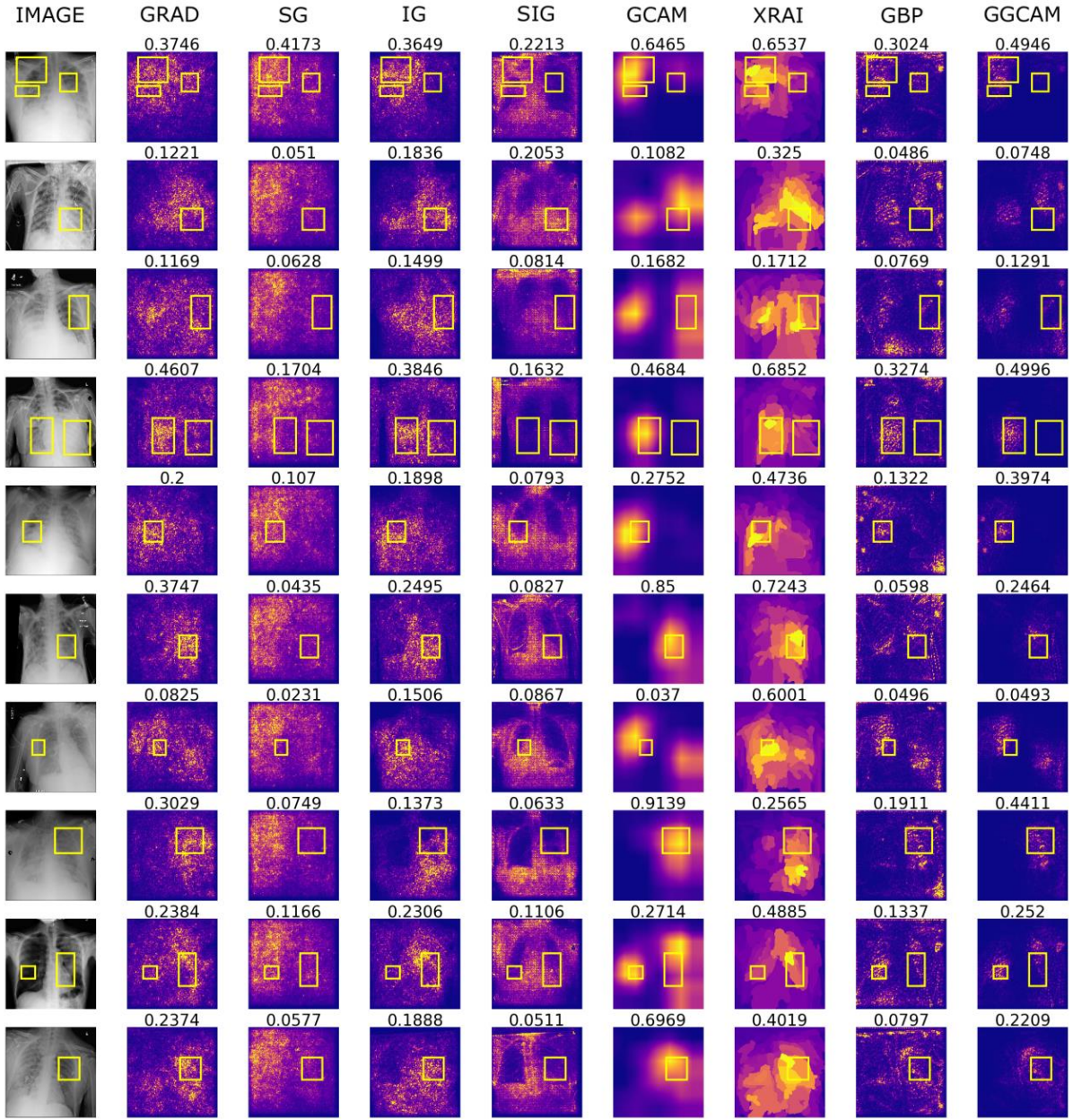

**Fig. S1:** Additional visualizations of saliency maps generated using an InceptionV3 on examples from the RSNA Pneumonia dataset (with yellow bounding boxes). Numbers above each image denote the SSIM score.

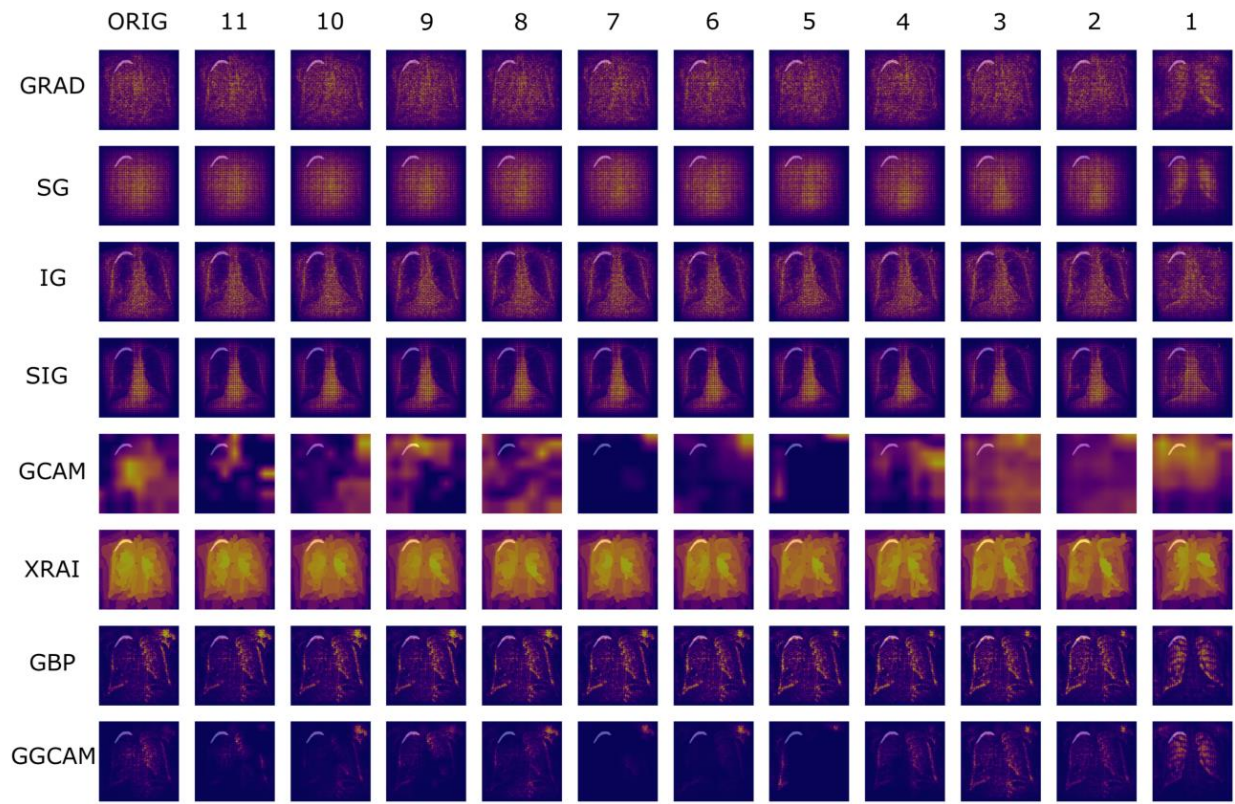

**Fig. S2:** Additional visualizations of cascading randomization using InceptionV3 on examples from the SIIM Pneumothorax dataset (with transparent segmentations).

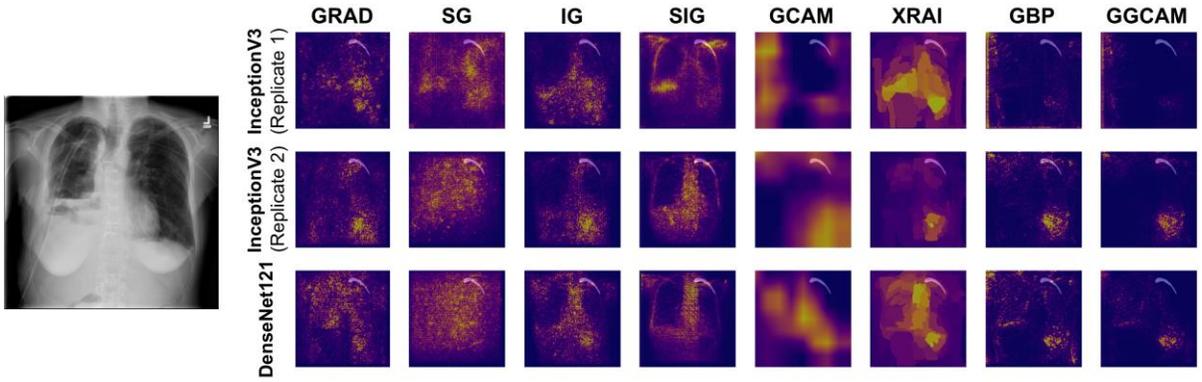

(a)

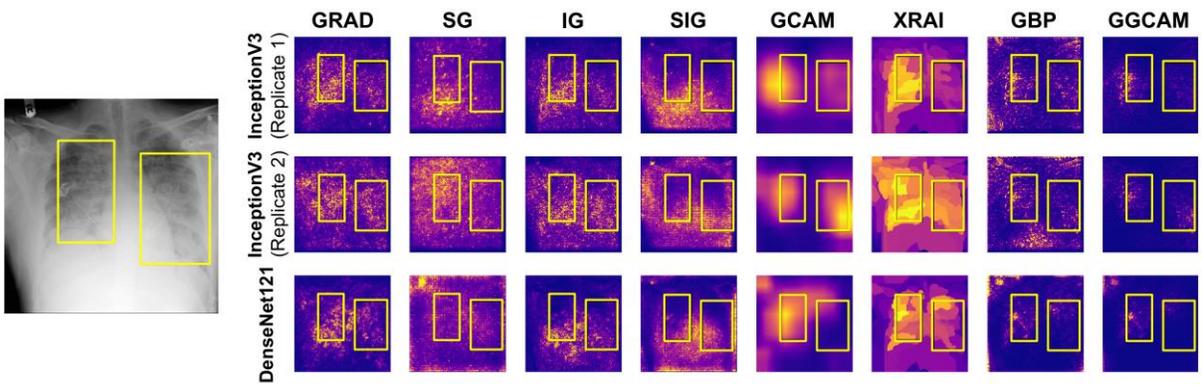

(b)

**Fig. S3:** Two additional examples of repeatability (InceptionV3 Replicates 1 and 2) and reproducibility (InceptionV3 and DenseNet121) for the (a) SIIM-ACR Pneumothorax dataset with transparent segmentations and (b) RSNA Pneumonia dataset with yellow bounding boxes

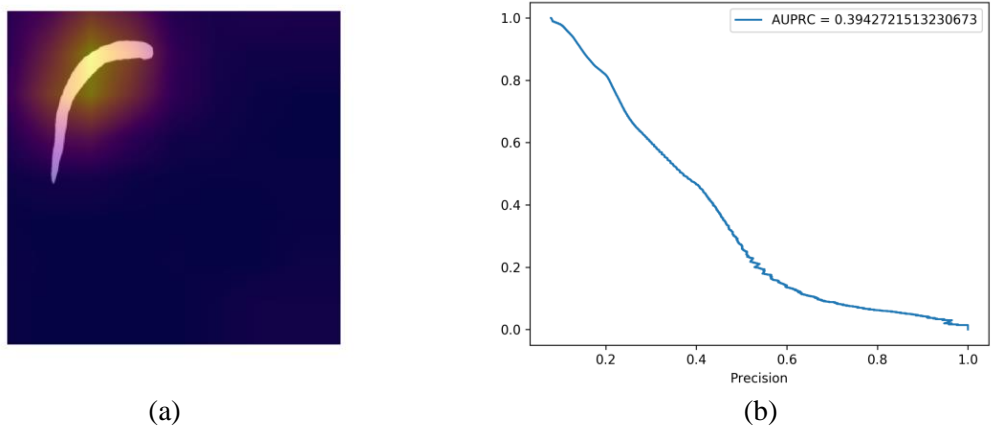

(a)                            (b)

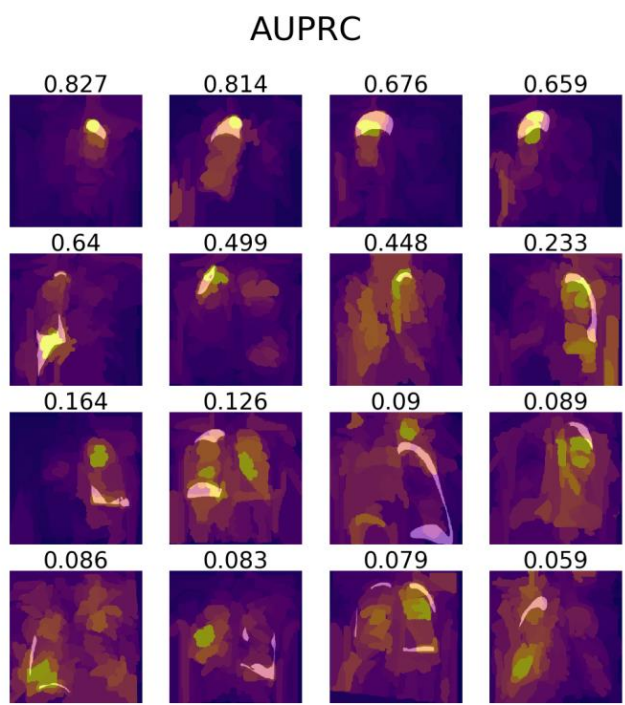

(c)

**Fig. S4:** (a) An example of a transparent pneumothorax segmentation compared to its corresponding GradCAM saliency map; (b) the corresponding precision-recall curve for the saliency map's overlap with the ground truth segmentation; (c) examples of XRAI saliency map overlap with ground truth pneumothorax segmentation with corresponding value of AUPRC.

**Supplemental Tables**

**Table S1:** Mean Structural SIMilarity (SSIM) index scores of saliency maps for fully randomized models (i.e., after cascading randomization has reached the bottom-most layer) and degradation thresholds (defined in the Results section) for the SIIM-ACR Pneumothorax segmentation and RSNA Pneumonia detection datasets. GRAD-Gradient Explanation; SG-Smooth Gradients; IG-Integrated Gradients; SIG-Smooth Integrated Gradients; GCAM-GradCAM; GBP-Guided-backprop; GGCAM-Guided GradCAM

|  | SSIM of Fully Randomized Model | | Degradation Threshold | |
| --- | --- | --- | --- | --- |
|  | Segmentation | Detection | Segmentation | Detection |
| **GRAD** | 0.250 ± 0.046 | 0.173 ± 0.001 | 0.239 ± 0.046 | 0.200 ± 0.029 |
| **SG** | 0.565 ± 0.204 | 0.253 ± 0.024 | 0.314 ± 0.204 | 0.315 ± 0.010 |
| **IG** | 0.306 ± 0.043 | 0.230 ± 0.001 | 0.239 ± 0.043 | 0.212 ± 0.036 |
| **SIG** | 0.567 ± 0.134 | 0.324 ± 0.013 | 0.501 ± 0.134 | 0.600 ± 0.114 |
| **GCAM** | 0.430 ± 0.106 | 0.369 ± 0.103 | 0.435 ± 0.106 | 0.542 ± 0.137 |
| **XRAI** | 0.814 ± 0.095 | 0.723 ± 0.005 | 0.631 ± 0.095 | 0.646 ± 0.053 |
| **GBP** | 0.355 ± 0.076 | 0.234 ± 0.022 | 0.359 ± 0.076 | 0.299 ± 0.079 |
| **GGCAM** | 0.409 ± 0.062 | 0.494 ± 0.029 | 0.627 ± 0.062 | 0.564 ± 0.092 |

**Table S2:** Mean AUROC (Area under Reciever Operating Characteristics) values for the SIIM-ACR Pneumothorax segmentation and RSNA Pneumonia detection datasets. GRAD-Gradient Explanation; SG-Smooth Gradients; IG-Integrated Gradients; SIG-Smooth Integrated Gradients; GCAM-GradCAM; GBP-Guided-backprop; GGCAM-Guided GradCAM; AVG - Average Mask; BASE - Baseline (RetinaNet/UNet)

|       | **Segmentation**   | **Detection**      |
|-------|--------------------|--------------------|
| **GRAD**  | $0.697 \pm 0.131$ | $0.787 \pm 0.731$ |
| **SD**    | $0.699 \pm 0.116$ | $0.624 \pm 0.102$ |
| **IG**    | $0.669 \pm 0.137$ | $0.787 \pm 0.069$ |
| **SIG**   | $0.640 \pm 0.150$ | $0.608 \pm 0.115$ |
| **GCAM**  | $0.651 \pm 0.277$ | $0.809 \pm 0.163$ |
| **XRAI**  | $0.800 \pm 0.164$ | $0.890 \pm 0.076$ |
| **GBP**   | $0.707 \pm 0.134$ | $0.622 \pm 0.109$ |
| **GGCAM** | $0.687 \pm 0.252$ | $0.763 \pm 0.143$ |
| **AVG**   | $0.730 \pm 0.191$ | $0.886 \pm 0.084$ |
| **BASE**  | $0.874 \pm 0.117$ | $0.947 \pm 0.040$ |